\relax
\documentclass[twocolumn]{article}
\usepackage{cite}
\usepackage{times}  
\usepackage{helvet}  
\usepackage{courier}  
\usepackage{url}  
\usepackage{graphicx}  
\usepackage[cmex10]{amsmath}
\usepackage{multirow}
\usepackage{paralist}
\usepackage{diagbox}
\usepackage{framed}
\usepackage{color}
\usepackage{tikz}
\usepackage{listings}
\usepackage{threeparttable}

\definecolor{dkgreen}{rgb}{0,0.6,0}
\definecolor{gray}{rgb}{0.5,0.5,0.5}
\definecolor{mauve}{rgb}{0.58,0,0.82}

\begin{document}
%
\title{Automatic Generation of Text Descriptive Comments for Code Blocks}
\author{
		Yuding Liang and
		Kenny Q. Zhu$^*$
		\\[0.5ex]
		\url{liangyuding@sjtu.edu.cn, kzhu@cs.sjtu.edu.cn}\\[0.5ex]
		Department of Computer Science and Engineering\\[0.5ex]
		Shanghai Jiao Tong University\\[0.5ex]
		800 Dongchuan Road, Shanghai, China 200240
}

\maketitle
\begin{abstract}
We propose a framework to automatically generate descriptive
comments for source code blocks. While this problem has been studied by many researchers previously, their methods are mostly based on
fixed template and achieves poor results.
Our framework does not rely on any template, but makes use
of a new recursive neural network called Code-RNN to extract features
from the source code and embed them into one vector. When this vector
representation is input to a new recurrent neural network (Code-GRU),
the overall framework generates text descriptions of the code with
accuracy (Rouge-2 value) significantly higher than other learning-based
approaches such as sequence-to-sequence model. The Code-RNN
model can also be used in other scenario where the representation
of code is required.~\footnote{Kenny Q. Zhu is the contact author.}
\end{abstract}

\section{Introduction}

Real-world software development involves large source code repositories.
Reading and trying to understand other people's code in such repositories
is a difficult and unpleasant process for many software developers,
especially when the code is not sufficiently commented.
For example, if the Java method in Fig.~\ref{figure:sourceCodeExample} does not
have the comment in the beginning, it will take the programmer
quite some efforts to grasp the meaning of the code.
However, with a meaningful sentence such as
``calculates dot product of two points''
as a descriptive comment, programmer's
productivity can be tremendously improved.


\begin{figure}[th]
\begin{lstlisting}
    /* Calculates dot product of two points.
     * @return float */
    public static float ccpDot(final CGPoint v1, final CGPoint v2) {
        return v1.x * v2.x + v1.y * v2.y;
    }
\end{lstlisting}
\caption{\label{figure:sourceCodeExample}source code example}
\end{figure}

A related scenario happens when one wants to search for a piece of code with
a specific functionality or meaning. Ordinary keyword search would not work
because expressions in programs can be quite different from natural languages.
If methods are annotated with meaningful natural language comments, then
keyword matching or even fuzzy semantic search can be achieved.

Even though comments are so useful, programmers are not using them enough in their
coding. Table \ref{table:repos} shows the number of methods in
ten actively developed Java repositories from Github, and those of
which annotated with a descriptive comment.
On average, only 15.4\% of the methods are commented.


\begin{table*}[th]
\caption{\label{table:repos} Ten Active Projects on Github}
\center
\scriptsize{
\begin{tabular}{|c|c|c|c|c|c|}
\hline
Project & Description & \# of bytes & \# of Java Files& \# of Methods& \# Methods Commented\\
\hline \hline
Activiti & a light-weight workflow and Business Process Management (BPM) Platform & 168M & 2939 & 15875 & 1080 \\
\hline
aima-java & Java implementation of algorithms from ``Artificial Intelligence - A Modern Approach'' & 182M & 889 & 4078 & 1130\\
\hline
neo4j & the world¡¯s leading Graph Database. & 270M & 4125 & 24529 & 1197 \\
\hline
cocos2d & cocos2d for android, based on cocos2d-android-0.82 & 78M & 512 & 3677 & 1182 \\
\hline
rhino & a Java implementation of JavaScript.& 21M & 352 & 4610 & 1195 \\
\hline
spring-batch & a framework for writing offline and batch applications using Spring and Java & 56M & 1742 & 7936 & 1827 \\
\hline
Smack & an open source, highly modular, easy to use, XMPP client library written in Java & 41M & 1335 & 5034 & 2344\\
\hline
guava & Java-based projects: collections, caching, primitives & 80M & 1710 & 20321 & 3079 \\
\hline
jersey & a REST framework that provides JAX-RS Reference Implementation and more. & 73M & 2743 & 14374 & 2540\\
\hline
libgdx & a cross-platform Java game development framework based on OpenGL (ES)& 989M & 1906 & 18889 & 2828 \\
\hline
\end{tabular}

\begin{tablenotes}
 \item[1] A comment here refers to the description at the beginning
of a method, with more than eight words.
\end{tablenotes}
}
\end{table*}

To automatically generate descriptive comments from source code,
one needs a way of accurately representing the semantics of code blocks.
One potential solution is to treat each code block as a document
and represent it by a topic distribution using models such as LDA~\cite{blei2003latent}.
However, topic models, when applied to source code,
have several limitations:

\begin{itemize}
\item a topic model treats documents as a bag of words and ignores the
structural information such as programming language syntax and
function or method calls in the code;
\item the contribution of lexical semantics to the meaning of code
is exaggerated;
\item comments produced can only be words but not phrases or sentences.
\end{itemize}

%

One step toward generating readable comments is to
use templates~\cite{mcburney2014automatic,sridhara2010towards}.
The disadvantage is that comments created by templates are often
very similar to each other and only relevant to parts of the code
that fit the template.
For example, the comment generated by McBurney's model
for Fig.~\ref{figure:sourceCodeExample} is fairly useless:
{\em ``This method handles the ccp dot and returns a float.
ccpDot() seems less important than average because it
is not called by any methods.''
}

\begin{figure}[th]
	\centering
	\includegraphics[width=\linewidth]{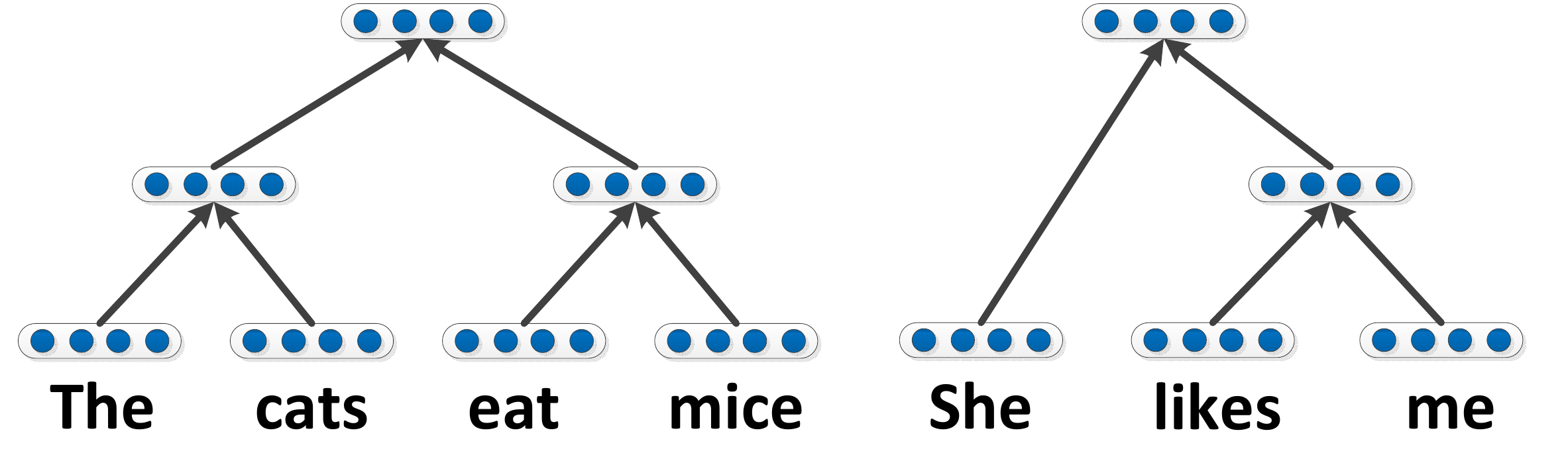}
	\caption{The Recursive Neural Networks of Two Sentences}
	\label{rnn:nlp}
\end{figure}

To overcome these problems,
in this paper, we propose to use Recursive Neural Network
(RNN)~\cite{socher2011parsing,socher2011semi} to
combine the semantic and structural information from code.
Recursive NN has previously been applied to parse trees of natural language sentences, such
as the example of two sentences in Fig.~\ref{rnn:nlp}.
In our problem, source codes can be
accurately parsed into their parse trees, so recursive NN can be applied
in our work readily. To this end, we design a new recursive NN called
Code-RNN to extract the features from the source code.

Using Code-RNN to train from the source code, we can get
a vector representation of each code block and this vector contains
rich semantics of the code block, just like word
vectors~\cite{mikolov2013distributed}. We then use a Recurrent Neural
Network to learn to generate meaningful comments.
Existing recurrent NN 
does not take good advantage of the code block representation vectors.
Thus we propose a new GRU~\cite{cho2014properties} cell that does a better job.

In sum, this paper makes the following contributions:
\begin{itemize}
\item by designing a new Recursive Neural Network, \emph{Code-RNN},
we are able to describe the {\em structural information} of source code;
\item with the new design of a GRU cell, namely \emph{Code-GRU}, we make the best
out of code block representation vector to effectively generate comments
for source codes;
\item the overall framework achieves remarkable accuracy (Rouge-2 value) in the task of
generating descriptive comments for Java methods,
compared to state-of-the-art approaches.
\end{itemize}


\section{Framework}

In this section, we introduce the details of how to represent source code and how to use the representation vector of source code to generate comments.

\begin{figure*}[th]
\begin{minipage}{0.15\linewidth}
  \begin{lstlisting}

  if (!found){
    allFound = false;
  }
  if (allFound){
    return true;
  }
\end{lstlisting}
 \centerline{source code}
\end{minipage}
\hfill
\begin{minipage}{0.9\linewidth}
  \centering
	\includegraphics[width=0.9\linewidth]{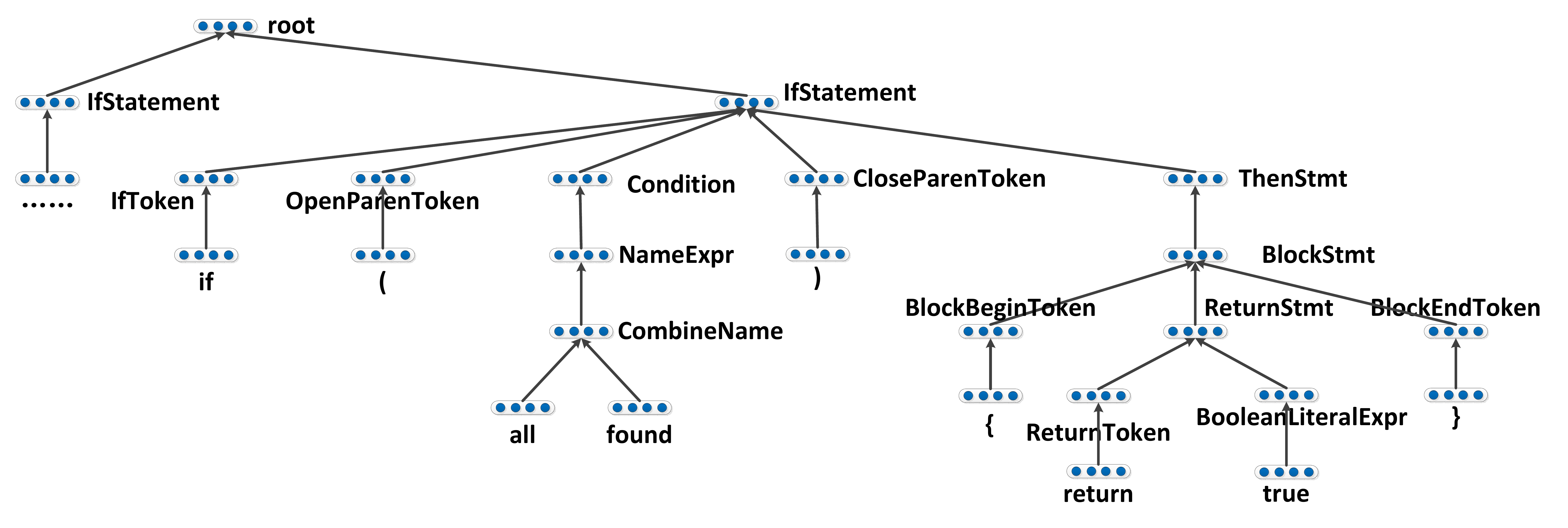}
 \centerline{Code-RNN}
\end{minipage}

\caption{Code-RNN Example}\label{fig:code_rnn}
\end{figure*}

\subsection{Code Representation}

We propose a new kind of recursive neural network called Code-RNN to encapsulate
the critical structural information of the source code.
Code-RNN is an arbitrary tree form while other recursive neural nets
used in NLP are typically binary trees. Fig. \ref{fig:code_rnn} shows an
example of Code-RNN for a small piece of Java code.

In Code-RNN, every parse tree of a program is encoded into a neural network,
where the structure of the network is exactly the parse tree itself and
each syntactic node in the
parse tree is represented by a vector representation.~\footnote{We use
JavaParser from \url{https://github.com/javaparser/javaparser} to
generate parse tree for Java code in this paper.}
One unique internal node ``CombineName'' indicates a compound identifier that is
the concatenation of several primitive words, for example, ``allFound'' can be split into
``all'' and ``found''. More on the semantics of identifier will be discussed later in
this section.


There are two models for the Code-RNN, namely \emph{Sum Model} and \emph{Average Model}:
\begin{enumerate}
\item{Sum Model}
\begin{equation}\label{eq:code_rnn_sum}
V = V_{node} + f(\textbf{W} \times \sum_{c \in C}{ V_{c}} + \textbf{b})
\end{equation}

\item{Average Model}
\begin{equation}\label{eq:code_rnn_average}
V = V_{node} + f(\textbf{W} \times \frac{1}{n}\sum_{c \in C}{V_{c}} + \textbf{b})
\end{equation}
\end{enumerate}
Here $V$ is the vector representation of sub-tree rooted at $N$;
$V_{node}$ is the vector that represents the syntactic type of $N$ itself, e.g.,
{\em IfStatement};
$C$ is the set of all child nodes of $N$; $V_{c}$ is the vector that represents
a subtree rooted at $c$, one of $N$'s children.
During the training, $W$ and $b$ are tuned.
$V$, $V_{node}$ and $V_{c}$ are calculated based on the structure
of neural network. $f$ is \emph{RELU} activation function.


These equations are applied recursively, bottom-up through the Code-RNN at
every internal node, to obtain the vector representation of the root node,
which is also the vector of the entire code piece.

\subsubsection{Identifier Semantics}
\label{sec:identifier}

In this work, we adopt two ways to extract the semantics from the identifiers.
One is to split all the long forms to multiple words and
the other one is to recover the full words from abbreviations.

Table \ref{table:splitID} shows some example identifiers and the results
of splitting. Many identifiers in the source code are combination of English words,
with the first letter of the word in upper case, or joined together using
underscores. We thus define simple rules to extract the original
English words accordingly. These words are further connected by the
``CombineName'' node in the code-RNN.

Table \ref{table:abbr} shows some abbreviations and
their intended meaning. We can infer the full-versions by
looking for longer forms in the context of the identifier in the code.
Specifically, we compare the identifier with the word list generated from
the context of the identifier to see whether the identifier's name
is a substring of some word from the list, or is the combination of
the initial of the words in the list.
If the list contains only one word, we just check if the identifier
is part of that word. If so, we conclude that the identifier is the
abbreviation of that word with higher probability.
If the list contains multiple words, we can collect all the initials
of the words in the list to see whether the identifier is part of
this collection. Suppose the code fragment is
\begin{lstlisting}
Matrix dm = new DoubleMatrix(confusionMatrix);
\end{lstlisting}
We search for the original words of ``dm'' as follows.
Since ``dm'' is not the substring of any word in the context,
we collect the initials of the contextual words in a list:
``m'' ``dm'' and ``cm''.
Therefore, ``dm'' is an abbreviation of ``DoubleMatrix''.

\begin{table}[th]
\caption{\label{table:splitID} Example of Split Identifiers}
\center
\scriptsize{
\begin{tabular}{|c|c|}
\hline
Identifier & Words \\
\hline \hline
contextInitialize & context, initialize\\
\hline
apiSettings & api, settings\\
\hline
buildDataDictionary & build, data, dictionary\\
\hline
add\_result & add, result\\
\hline
\end{tabular}
}
\end{table}

\begin{table}[th]
\caption{Example of Abbreviation}
\label{table:abbr}
\center
\scriptsize{
\begin{tabular}{|c|c|c|}
\hline
Abbreviation & Origin & Context\\
\hline \hline
val & value & key.value()\\
\hline
cm & confusion, matrix & new ConfusionMatrix()\\
\hline
conf & configuration & context.getConfiguration()\\
\hline
rnd & random & RandomUtils.getRandom()\\
\hline
\end{tabular}
}
\end{table}

\subsubsection{Training}
Each source code block in the training data has a class label.
Our objective function is:
\begin{equation}
\small{
\mathop{\arg\min} CrossEntropy(softmax(W_{s}V_{m} + b_{s}),V_{label})\label{eq:object}
}
\end{equation}
where $V_{m}$ is the representation vector of source code, $V_{label}$ is
an one-hot vector to represent the class label.
$W_{s}$ and $b_{s}$ are parameters for softmax function and will be
tuned during training.  We use AdaGrad~\cite{duchi2011adaptive} to apply
unique learning rate to each parameter.

\subsection{Comment Generation}
\begin{figure}[th]
\centering
	\includegraphics[width=0.8\linewidth]{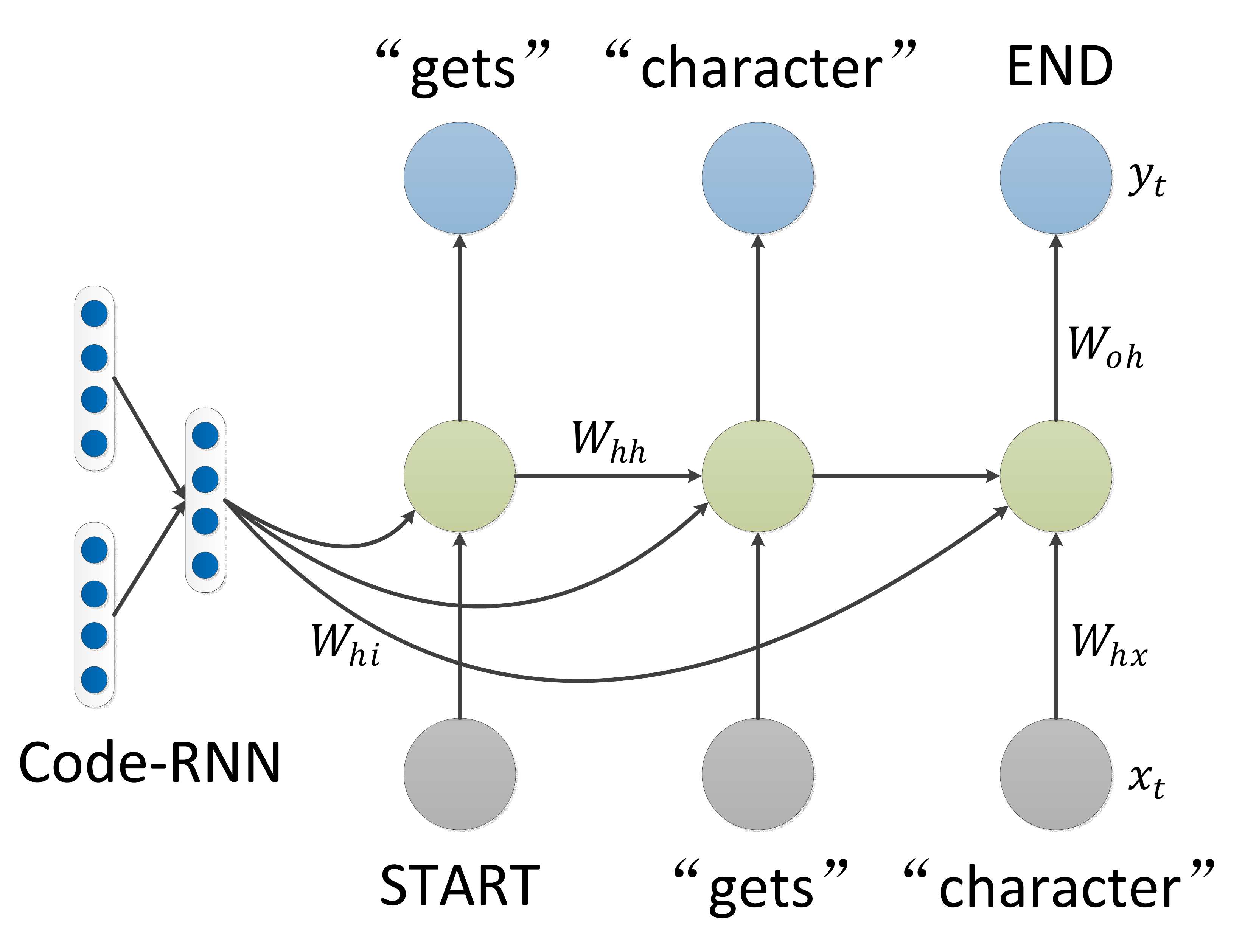}
\caption{Comment Generation}\label{fig:comment_generate}
\end{figure}

Existing work~\cite{elman1990finding,sutskever2011generating,mikolov2010recurrent} has used Recurrent Neural Network to generate sentences.
However, one challenge to utilize the code block representation vector
in Recurrent NN is that we can not feed the code block representation vector to
the Recurrent NN cell directly.
We thus propose a variation of the GRU based RNN.
Fig. \ref{fig:comment_generate} shows our comment generation process.

We use pre-trained model Code-RNN to get the representation vector of
the input code block $V_{m}$. This vector $V_{m}$ is fixed during training of
comment generation model. Then we feed code block vector into the
RNN (Recurrent Neural Network) model at every step.
For example in Fig.~\ref{fig:comment_generate}, we input the START token
as the initial input of model and feed the code block vector into the hidden
layer. After calculating the output of this step, we do the
back-propagation. Then at step two, we input the word ``gets'' and
feed the code block vector $V_{m}$ into hidden layer again,
and receive the $h_{t-1}$ from the step one. We repeat the above
process to tune all parameters. The equations of comment generation model
are listed below.

\begin{align}
z_{t} &= \sigma(W_{z}\cdot[h_{t-1},x_{t}])\\
r_{t} &= \sigma(W_{r}\cdot[h_{t-1},x_{t}])\\
c_{t} &= \sigma(W_{c}\cdot[h_{t-1},x_{t}]) \label{eq:ct} \\
\tilde{h_{t}} &= tanh(W\cdot[r_{t}*h_{t-1},c_{t}*V_{m},x_{t}]) \label{eq:ct2}\\
h_{t} &= (1-z_{t})*h_{t-1} + z_{t}*\tilde{h_{t}} \\
y_{t} &= softmax(W_{oh}h_{t} + b_{o})
\end{align}
where $V_{m}$ is the code block representation vector, $h_{t-1}$ is the previous state and $x_{t}$ is the input word of this step.

\begin{figure}[th]
\centering
	\includegraphics[width=0.8\linewidth]{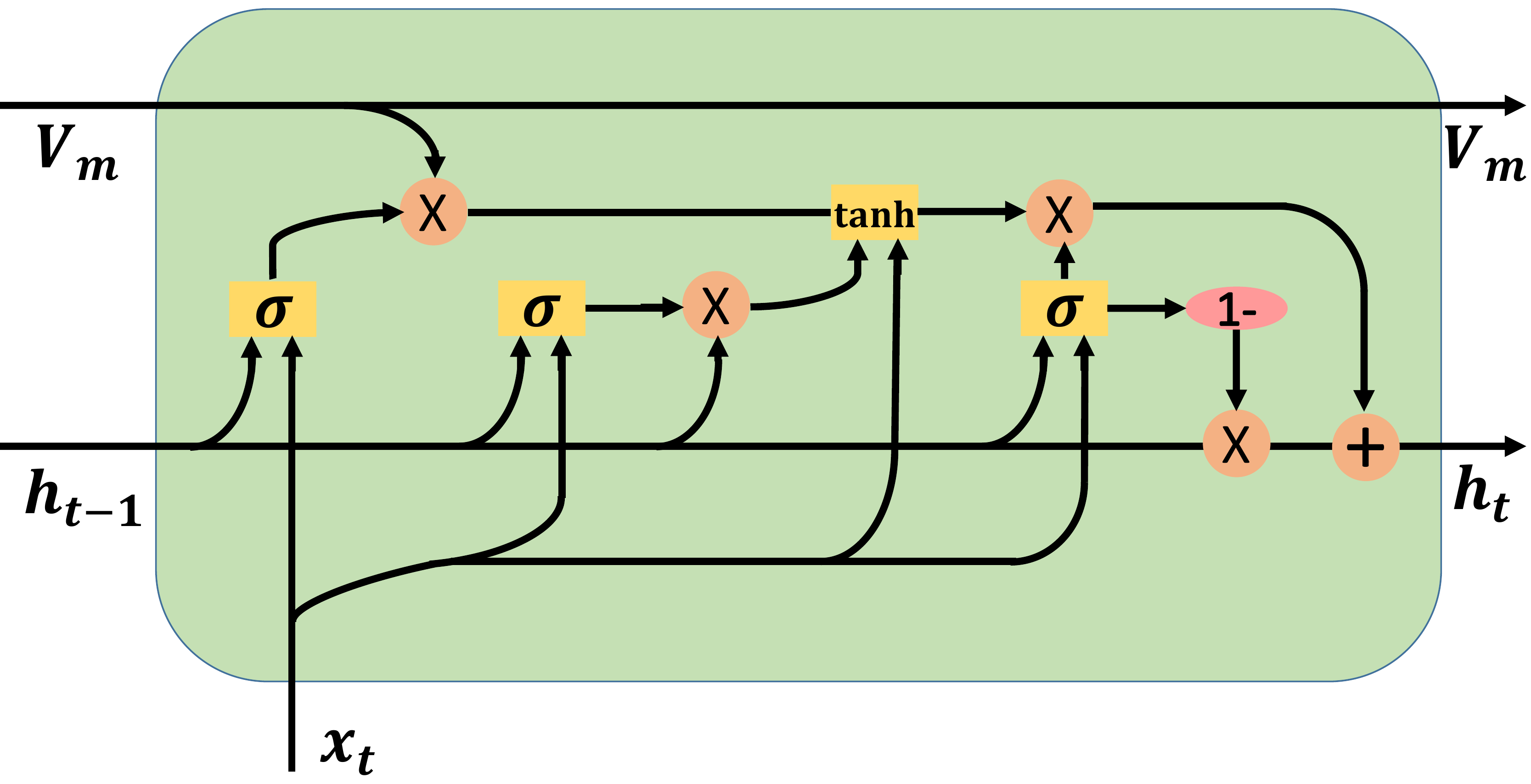}
\caption{Structure of Code-GRU}\label{fig:new_gru}
\end{figure}

To better use the code block vectors, our model differs from existing RNNs,
particularly in the definition of $c_{t}$ in the Equation \ref{eq:ct}
and \ref{eq:ct2}.
The new RNN cell, illustrated in Fig. \ref{fig:new_gru},
aims to strengthen the effect of code block vectors.
This modified GRU is hereinafter called \emph{Code-GRU}.
Code block vector contains all information of code block but not all information
is useful at all steps. Therefore, we add a new gate called choose gate
to determine which dimension of code block vector would work in Code-GRU.
In Fig \ref{fig:new_gru}, the left gate is the choose gate,
and the other two gates are the same as the original GRU.

During test time, we input the ``START'' token at first and choose the
most probable word as the output. Then from the second step the input words of every step are the output words of previous one step until the output is ``END'' token. So that we can get an automatically generated comment for code blocks
in our model.

To gain better results, we also apply the beam search while testing.
We adopt a variant of beam search with a length penalty described
in~\cite{wu2016google}. In this beam search model, there are two parameters:
beam size and weight for the length penalty. We tune these
two parameters on the validation set to determine which values to use.
Our tuning ranges are:
\begin{itemize}
    \item beam size: [1, 2, 3, 4, 5, 6, 7, 8, 9, 10]
    \item weight for the length penalty: [0, 0.1, 0.2, 0.3, 0.4, 0.5, 0.6, 0.7, 0.8, 0.9, 1.0]
\end{itemize}

\section{Evaluation}
Our evaluation comes in two parts. In the first part, we evaluate Code-RNN
model's ability to classify different source code blocks into $k$ known
categories. In the second part, we show the effectiveness of our comment
generation model by comparing with several state-of-the-art approaches in
both quantitative and qualitative assessments. The source code of our
approach as well as all data set is available at \url{https://adapt.seiee.sjtu.edu.cn/CodeComment/}.

\subsection{Source Code Classification}
\subsubsection{Data Set}
The goal is to classify a given Java method (we only use the body block without name and parameters) into a predefined set of
classes depending on its functionality.
Our data set comes from the Google Code Jam contest
(2008$\sim$2016), which there are multiple problems, each associated
with a number of correct solutions contributed by
programmers.~\footnote{All solutions are available at \url{http://www.go-hero.net/jam/16}.} Each solution is a Java method. The set of solutions for
the same problem are considered to function identically and belong
to the same class in this work. We use the solutions (10,724 methods) of
6 problems as training set and the solutions (30 methods) of the other 6
problems as the test set. Notice that the problems in the training data
and the ones in the test data do not overlap. We specifically design the
data set this way because, many methods for the same problem tend to use
the same or similar set of identifiers, which is not true in real world
application.  The details of training set and test set are shown in
Table \ref{table:method_data}.

\renewcommand{\multirowsetup}{\centering}
\begin{table}[th]
\caption{Data Sets for Source Code Clustering}
\label{table:method_data}
\center
\scriptsize{
\begin{tabular}{|c|c|c|c|}
 \hline
  & Problem & Year &\# of methods  \\
 \hline \hline
 \multirow{6}{1.5cm}{Training Set} & Cookie Clicker Alpha & 2014 &
 1639\\
 \cline{2-4}
  & Counting Sheep & 2016 &
  1722\\
 \cline{2-4}
  & Magic Trick & 2014 &
  2234\\
 \cline{2-4}
 & Revenge of the Pancakes & 2016 &
 1214\\
 \cline{2-4}
 & Speaking in Tongues & 2012 &
 1689\\
 \cline{2-4}
 & Standing Ovation & 2015 &
 2226\\
 \hline
 \multirow{6}{1.5cm}{Test Set} & All Your Base & 2009 & 5 \\ 
 \cline{2-4}
 & Consonants & 2013 & 5 \\ 
 \cline{2-4}
 & Dijkstra & 2015 & 5 \\ 
 \cline{2-4}
 & GoroSort & 2011 & 5 \\
 \cline{2-4}
 & Osmos & 2013 & 5 \\ 
 \cline{2-4}
 & Part Elf & 2014 & 5 \\ 
 \hline
\end{tabular}
}
\end{table}

%

\subsubsection{Baselines}

We compare Code-RNN with two baseline approaches.
The first one is called language embedding (\emph{LE}) and only treats the
source code as a sequence of words, minus the special symbols
(e.g., ``\$'', ``('', ``+'', $\cdots$). All concatenated words are preprocessed
into primitive words as previously discussed. Then the whole code can be
represented by either the sum (\emph{LES}) or the average (\emph{LEA})  of word vectors
of this sequence, trained in this model.
This approach basically focuses on the word semantics only and ignores the
structural information from the source code.

The second baseline is a variant of Code-RNN, which preprocesses the code parse
tree by consistently replacing the identifier names with placeholders before
computing the overall representation of the tree. This variant focuses on
the structural properties only and ignores the word semantics.

\subsubsection{Result of Classification}
At test time, when a method is classified into a class label,
we need to determine which test problem this class label refers to.
To that end, we compute the accuracy of classification for
all possible class label assignment and use the highest
accuracy as the one given by a model.

Table \ref{table:purity} shows the purity of the produced classes, the F1
and accuracy of the 6-class classification problem by different methods. It is
clear that Code-RNN (avg) perform better uniformly
than the baselines that use only word semantics or only structural information.
Therefore, in the rest of this section, we will use Code-RNN(avg) model to
create vector representation for a given method to be used for
comment generation.
The F1 score for each individual problem is also included in
Table \ref{table:F1_30}.

\begin{table}[th]
\caption{Purity, Average F1 and Accuracy}
\label{table:purity}
\center
\scriptsize{
\begin{tabular}{lccc}
 \hline
 & Purity &  F1 & Accuracy \\ 
 \hline
 LEA & 0.400 &  0.3515  & 0.3667\\
 LES & 0.3667 & 0.2846 & 0.3667 \\
CRA(ni) &  0.4667 & 0.4167 & 0.4667\\
CRS(ni) &  0.4667 & 0.4187 & 0.4667 \\
 CRA & \textbf{0.533} &  \textbf{0.4774} & \textbf{0.5} \\
 CRS & 0.4667 & 0.3945 & 0.4333\\
 \hline

\end{tabular}

\begin{tablenotes}
  \item[1] LEA = Language Embedding Average model;
  LES = Language Embedding Sum model;
  CRA = Code-RNN Average model;
  CRS = Code-RNN Sum model; (ni) = no identifier.
\end{tablenotes}
}
\end{table}

\begin{table}[th]
\caption{F1 scores of individual problems}
\center
\scriptsize{
\label{table:F1_30}
\begin{tabular}{c@{\ \ \ \ }c@{\ \ \ \ }c@{\ \ \ \ }c@{\ \ \ \ }c@{\ \ \ \ }c@{\ \ \ \ }c@{\ \ \ \ }c}
 \hline
 &Dijkstra&Part Elf&All Your Base&GoroSort&Consonants&Osmos \\
 \hline
 LEA
 & 0.25 & {0.33} & 0.43 & {0.33} & 0.4 & 0.36 \\
 LES
 & 0.33 & 0 & 0.53 & 0 & 0.53 & 0.31 \\
 CRA(ni) & 0.6 & 0 & {0.44} & {0.40} & {0.56} & 0.5 \\

 CRS(ni) & {0.62} & 0.29 & {0.67} & {0.5} & 0.44 & 0 \\
 CRA
 & {0.67} & 0 & {0.6} & {0.57} & {0.53} & {0.52} \\
 CRS
 & {0.73} & {0} & 0.44 & 0.55 & 0.4 & 0.25 \\
\hline
\end{tabular}
}

\end{table}

\subsection{Comment Generation Model}
\subsubsection{Data Set}
We use ten open-source Java code repositories from GitHub for this
experiment (see Table \ref{table:repos}). In each of these repositories
we extract descriptive comment and the corresponding method pairs. Constructor
methods are excluded from this exercise.  These pairs are then used
for training and test. Notice that all the method names and parameters are excluded from
training and test.

\subsubsection{Baselines}
We compare our approach with four baseline methods.

\begin{itemize}
\item \emph{Moses}\footnote{Home page of Moses is \url{http://www.statmt.org/moses/}.}
 is a statistical machine translation system.
We regard the source codes as the source language and the comments as
the target, and use Moses to translate from the source to the target.
\item \emph{CODE-NN}~\cite{iyer2016summarizing} is the first model to use neural network to create sentences for source code. In this model author used LSTM and attention mechanism to generate sentences. The original data set for CODE-NN are
StackOverFlow thread title and code snippet pairs. \footnote{Data comes from
\url{https://stackoverflow.com/}. Source code of CODE-NN is available
from \url{https://github.com/sriniiyer/codenn}.}. In this experiment,
we use the comment-code pair data in place of the title-snippet data.
\item We apply the \emph{sequence-to-sequence (seq2seq)} model used
in machine translation \cite{Britz:2017} and treat the code as
a sequence of words and the comment as another sequence.
\item A. Karpathy and L. Fei-Fei~\cite{karpathy2015deep} proposed a meaningful method to generate image descriptions. It also used Recurrent NN and representation vector, so we apply this method to comment generation model. The main equations are:
\begin{align}
b_{v} &= W_{hi}V_{m}\\
h_{t} &= f(W_{hx}x_{t} + W_{hh}h_{t-1} + b_{h} + b_{v})\\
y_{t} &= softmax(W_{oh}h_{t} + b_{o})
\end{align}
where $W_{hi}$, $W_{hx}$, $W_{hh}$, $W_{oh}$, $x_{i}$ and $b_{h}$, $b_{o}$ are
parameters to be learned, and $V_{m}$ is the method vector.
We call this model \emph{Basic RNN}.
\end{itemize}
Moses and CODE-NN has its own terminate condition. Seq2Seq, Basic RNN and
our model run 800 epochs during training time.
For one project, we separate the commented methods into three parts:
training set, validation set and test set. We tune the hyper parameter on
the validation set. The results of ten repositories are shown
in Table \ref{table:rouge}.

\subsubsection{Evaluation Metric}
We evaluate the quality of comment generation by the Rouge
method\cite{lin2004rouge}.
Rouge model counts the number of overlapping units between
generated sentence and target sentence.
We choose Rouge-2 score in this paper where word based 2-grams are used as
the unit, as it is the most commonly used in evaluating automatic text
generation such as summarization.

\begin{table}[th]
\caption{Rouge-2 Values for Different Methods}
\center
\scriptsize{
 \begin{tabular}{@{\ }p{0.08\columnwidth}@{\ \ \ \ }c@{\ \ }c@{\ \ }c@{\ \ }p{0.065\columnwidth}@{\ \ }c@{\ \ }c@{\ \ }c@{\ \ }p{0.08\columnwidth}@{\ \ }c@{\ \ }c@{\ }}
 \hline
  & neo4j &cocos2d & jersey & aima-java & guava & Smack & Activiti & spring-batch & libgdx & rhino\\
 \hline
MOSES
& 0.076 & 0.147 & 0.081 & 0.144 & 0.134 & 0.145 & 0.104 & 0.147 & 0.212 & 0.082\\
\hline
CODE-NN
& 0.077 &{0.136} & 0.105 & 0.124 &0.153 & 0.135 & 0.103 & 0.184 & 0.208 & \textbf{0.171}\\
\hline
Seq2seq
& 0.039 & 0.115 & 0.183 & 0.108 & 0.152 & 0.109 & 0.158 &0.171  &  \textbf{0.247} & 0.169 \\ \hline
Basic RNN*
& 0.133 & 0.152 & 0.214 & {0.207} & 0.156 & 0.150 & \textbf{0.203} & \textbf{0.237} & 0.218 & 0.163\\
\hline
Code-GRU*
& \textbf{0.141} & \textbf{0.158} & \textbf{0.230} & \textbf{0.209} & \textbf{0.164} & \textbf{0.162} & 0.200 & 0.213 &  0.233 & 0.165\\
\hline
 \end{tabular}
 \begin{tablenotes}
  \item[1] *: both models use the method representation vector from Code-RNN.
\end{tablenotes}
}

\label{table:rouge}
\end{table}

\begin{figure*}[!htb]
\centering
\scriptsize{
\rule{\linewidth}{0.3mm}
\\[0.2cm]
\begin{minipage}{0.48\linewidth}
\centering
\hspace{.2cm}
\begin{lstlisting}
Project: jersey
    public long tryConvertToReadLock(long stamp) {
        long a = stamp & ABITS, m, s, next; WNode h;
        while (((s = state) & SBITS) == (stamp & SBITS)) {
            if ((m = s & ABITS) == 0L) {
                if (a != 0L)   break;
                else if (m < RFULL) {
                    if (U.compareAndSwapLong(this, STATE, s, next = s + RUNIT))
                        return next;}
                else if ((next = tryIncReaderOverflow(s)) != 0L) return next;}
            else if (m == WBIT) {
                if (a != m)     break;
                state = next = s + (WBIT + RUNIT);
                if ((h = whead) != null && h.status != 0)
                    release(h);
                return next;}
            else if (a != 0L && a < WBIT)   return stamp;
            else  break;}
        return 0L;}
\end{lstlisting}
\end{minipage}
 \hfill
\begin{minipage}{0.48\linewidth}
\begin{tabular}{|c|p{0.75\columnwidth}|}
 \hline
   {Gold} & if the lock state matches the given stamp performs one of the following actions if the stamp represents holding a write lock releases it and UNK a read lock or if a read lock returns it or if an optimistic read acquires a read lock and returns a read stamp only if immediately available this method returns zero in all other cases\\
 \hline
   {MOSES} & if the lock state matches the given if the lock state matches the given gets of processing sbits state matches the given sbits string of the lock hold abits l l break component of rfull that runs sets of processing of runit or create a new pattern if this inc reader overflow l create a human readable description of component of wbit break if the lock of processing of wbit runit h whead by the given status release the given action if the sum associated with the given component l lock state matches the given action wbit get returns break l
   \\
 \hline
   {CODE-NN} & returns code true if the lock is not a link org glassfish jersey server mvc\\
 \hline
   {Seq2Seq} & UNK a new item to the list of superclass timeout version\\
 \hline
   {Basic RNN*} & get a UNK to a link javax ws rs core UNK\\
 \hline
  {Code-GRU*} & if the lock state matches the given stamp performs one of the following actions if the stamp represents holding a write lock returns it or if a read lock if the write lock is available releases the read lock and returns a write stamp or if an optimistic read returns\\
 \hline
 \end{tabular}
 \end{minipage}
 \\[0.2cm]
 \rule{\linewidth}{0.3mm}
 \\[0.2cm]
\vfill
 \begin{minipage}{0.48\linewidth}
 \hspace{.2cm}
\begin{lstlisting}
project: cocos2d
    public static float ccpDot(final CGPoint v1, final CGPoint v2) {
        return v1.x * v2.x + v1.y * v2.y;    }
\end{lstlisting}
\end{minipage}
\hfill
\begin{minipage}{0.48\linewidth}
\begin{tabular}{|c|p{0.75\columnwidth}|}
 \hline
   {Gold} & Calculates dot product of two points \\
 \hline
   {MOSES} & subtract another subtract another the given vector \\
 \hline
   {CODE-NN} & rotates two points \\
 \hline
   {Seq2Seq} & returns the closest long to the specified value \\
 \hline
   {Basic RNN*} & calculates cross product of two points \\
 \hline
   {Code-GRU*} & calculates cross product of two points \\
 \hline
 \end{tabular}
 \end{minipage}
 \\[0.2cm]
\rule{\linewidth}{0.3mm}
\\[0.2cm]
\vfill
 \begin{minipage}{0.48\linewidth}
\hspace{.2cm}
\begin{lstlisting}
project: libgdx
    public static IntBuffer allocate (int capacity) {
      if (capacity < 0) {
        throw new IllegalArgumentException();
      }
      return BufferFactory.newIntBuffer(capacity);}
\end{lstlisting}
\end{minipage}
\hfill
\begin{minipage}{0.48\linewidth}
\begin{tabular}{|c|p{0.75\columnwidth}|}
 \hline
   {Gold} & creates an int buffer based on a newly allocated int array\\
 \hline
   {MOSES} & based on the creates a new backing buffer\\
 \hline
   {CODE-NN} & creates a byte buffer based on a newly allocated char array\\
 \hline
   {Seq2Seq} & creates a float buffer based on a newly allocated float array \\
 \hline
   {Basic RNN*} & creates a char buffer based on a newly allocated char array\\
 \hline
   {Code-GRU*} & creates a long buffer based on a newly allocated long array\\
 \hline
 \end{tabular}
 \end{minipage}
 \\[0.2cm]
 \rule{\linewidth}{0.3mm}

}
\caption{Examples of generated comments and corresponding code snippets}
\label{fig:comment_example}
\end{figure*}

\subsubsection{Examples of Generated Comment}
Fig. \ref{fig:comment_example} shows the comments generated by
the competing methods for three example Java methods coming from
different repositories.
Because we delete all punctuation from the training data,
the generated comments are without punctuation. Nonetheless, we can see that
comments by our Code-GRU model are generally more readable and meaningful.

In the first example, we can see that CODE-NN, Seq2Seq and Basic RNN's results are
poor and have almost nothing to do  with the Gold comment.
Even though both MOSES produces a sequence of words that look similar to the Gold
in the beginning, the rest of the result is less readable and does not have
any useful information. For example, ``if the lock state matches the given'' is output
repeatedly. MOSES also produces strange terms such as ``wbit'' and ``runit'' just
because they appeared in the source code.
In the contrast, Code-GRU's result is more readable and meaningful.


In the second example, there is not any useful word in the method body
so the results of MOSES, CODE-NN and Seq2Seq are bad.
Code-RNN can extract the structural information of source code and embed it
into a vector, so both models that use this vector, namely Basic RNN and
Code-GRU, can generate the relevant comments.

In the third example, although all results
change the type of the value, that is, Basic RNN changes ``int'' to ``char''
while Code-GRU changes to ``long''.  ``long'' and ``int'' are both numerical
types while ``char'' is not. Thus Code-GRU is better than Basic RNN.
For the result of Seq2Seq, although ``float'' is also a numerical type,
it is for real numbers, and not integers.

%
%

\section{Related Work}


Mining of source code repositories becomes increasingly popular in
recent years. Existing work in source code mining include
code search, clone detection, software evolution,
models of software development processes, bug localization,
software bug prediction, code summarization and so on.
Our work can be categorized as code summarization and comment generation.

Sridhara et al.~\cite{sridhara2010towards} proposed an automatic
comment generator that identifies the content for the summary and
generates natural language text that summarizes the method¡¯s overall
actions based on some templates.
Moreno et al.~\cite{moreno2013automatic} also proposed a
template based method but it is used on summarizing Java classes.
McBurney and McMillan~\cite{mcburney2014automatic}  presented a novel approach
for automatically generating summaries of Java methods that summarize
the context surrounding a method, rather than details from the
internals of the method.
These summarization techniques \cite{murphy1996lightweight,sridhara2011generating,moreno2013automatic,haiduc2010use} work by selecting a
subset of the statements and keywords from the code, and
then including information from those statements and keywords in the
summary. To improve them, Rodeghero et al.~\cite{rodeghero2014improving}
presented an eye-tracking study of programmers
during source code summarization, a tool for selecting keywords
based on the findings of the eye-tracking study.

These models are invariably based on templates and careful selection
of fragments of the input source code. In contrast, our model is
based on learning and neural network. There are also some models
that apply learning methods to mine source code.

Movshovitz-Attias and Cohen~\cite{movshovitz2013natural}
predicted comments using topic models and n-grams.
Like source code summarization,
Allamanis et al.~\cite{allamanis2015suggesting} proposed a
continuous embedding model to suggest accurate method and class names.


Iyer et al.~\cite{iyer2016summarizing} proposed a new model called CODE-NN
that uses Long Short Term Memory (LSTM) networks with attention
to produce sentences that can describe C\# code snippets and SQL queries.
Iyer et al.'s work has strong performance on two tasks,
code summarization and code retrieval.
This work is very similar to our work,
in that we both use the Recurrent NN to generate sentences for source code.
What differs is that we propose a new type of Recurrent NN.
Adrian et al.~\cite{kuhn2007semantic} utilized the information of
identifier names and comments to mine topic of source code repositories.
Punyamurthula~\cite{punyamurthula2015dynamic} used call graphs to
extract the metadata and dependency information from the
source code and used this information to analyze the source code
and get its topics.


In other related domains of source code mining,
code search is a popular research direction.
Most search engines solve the problem by keyword extraction and
signature matching.  Maarek et al.~\cite{maarek1991information} used
keywords extracted from man pages written in natural language and
their work is an early example of approaches based on keywords.
Rollins and Wing~\cite{rollins1991specifications} proposed an
approach to find code with the signatures present in code.
Mitchell~\cite{mitchell2008hoogle} combined signature matching with
keyword matching. Then Garcia et al.~\cite{garcia2016semantic} focused
on querying for semantic characteristics of code and
proposed a new approach which combines semantic characteristics and
keyword matching.

Cai~\cite{cai2016code} proposed a method
for code parallelization through sequential code search.
That method can also be used for clone detection.
Williams and Hollingsworth~\cite{williams2005automatic}
described a method to use the source code change history of a
software project to drive and help to refine the search for bugs.
Adhiselvam et al.~\cite{adhiselvam2015enhanced} used MRTBA algorithm
to localize bug to help programmers debug. The method proposed in this paper
can also benefit natural language search for code fragments.


%
%
%


\section{Conclusion}

In this paper we introduce a new Recursive Neural Network called \emph{Code-RNN} to extract the topic or function of the source code. This new Recursive Neural Network is the parse tree of the source code and we go through all the tree from leaf nodes to root node to get the final representation vector. Then we use this vector to classify the source code into some classes according to the function, and classification results are acceptable.
We further propose a new kind of GRU called \emph{Code-GRU} to utilize the vector created from Code-RNN to generate comments. We apply Code-GRU to ten source code repositories and gain the best result in most projects.
This frame work can also be applied to other programming languages as long as we have access to the parse tree of the input program.


As future work, we can add call graphs into our model,
so that Code-RNN can contain invocation information and extract more topics
from source code.

\section*{Acknowledgement}
This work was supported by Oracle-SJTU Joint Research Scheme, NSFC Grant No. 91646205£¬71421002 and 61373031, and SJTU funding project 16JCCS08. Hongfei Hu
contributed to the identifier semantics part of the work.
\bibliographystyle{plain}
\bibliography{aaai}

\end{document}